\documentclass[letterpaper, 10 pt, conference]{ieeeconf}  

\IEEEoverridecommandlockouts                              

\usepackage{amsmath} 
\usepackage{amssymb}  
\usepackage{graphicx}
\usepackage{xcolor}

\usepackage{multirow}

\title{\LARGE \bf
DeepOFormer: Deep Operator Learning with Domain-informed Features for Fatigue Life Prediction}

\author{Chenyang Li$^{1}$, Tanmay Sunil Kapure$^{1}$, Prokash Chandra Roy$^{2}$,  Zhengtao Gan$^{3}$, Bo Shen$^{1}$ 
\thanks{*This research was supported by the U.S. Army Combat Capabilities Development Command Data \& Analysis Center (DEVCOM DAC) under Contract No. W911QX24D0021. The views and conclusions contained in this document are those of the authors and should not be interpreted as representing the official policies, either expressed or implied, of the U.S. Army or the U.S. Government.}
\thanks{$^{1}$ New Jersey Institute of Technology, Newark, NJ 07102}%
\thanks{$^{2}$ University of Texas at El Paso, El Paso, TX 79968}%
\thanks{$^{3}$ Arizona State University, Phoenix, AZ 85004}%
\thanks{Corresponding Author: Dr. Bo Shen, Email: {\tt\small bo.shen@njit.edu}}%
}

\begin{document}

\maketitle
\thispagestyle{empty}
\pagestyle{empty}

\begin{abstract}
Fatigue life characterizes the duration a material can function before failure under specific environmental conditions, and is traditionally assessed using stress-life (S-N) curves. While machine learning and deep learning offer promising results for fatigue life prediction, they face the overfitting challenge because of the small size of fatigue experimental data in specific materials. To address this challenge, we propose, DeepOFormer, by formulating S-N curve prediction as an operator learning problem. DeepOFormer improves the deep operator learning framework with a Transformer-based encoder and a mean L2 relative error loss function. We also consider Stüssi, Weibull, and Pascual and Meeker (PM) features as domain-informed features. These features are motivated by empirical fatigue models. To evaluate the performance of our DeepOFormer, we compare it with different deep learning models and XGBoost on a dataset with 54 S-N curves of aluminum alloys. With seven different aluminum alloys selected for testing, our DeepOFormer achieves an $R^2$ of 0.9515, a mean absolute error of 0.2080, and a mean relative error of 0.5077, significantly outperforming state-of-the-art deep/machine learning methods including DeepONet, TabTransformer, and XGBoost, etc. The results highlight that our Deep0Former integrating with domain-informed features substantially improves prediction accuracy and generalization capabilities for fatigue life prediction in aluminum alloys.  

\end{abstract}

\section{Introduction}
Fatigue life characterizes the duration over which a material can function before failure occurs under specific environmental conditions~\cite{tawaf2014fatigue,santecchia2016review}. The reliability and security of aluminum alloys are commonly evaluated by measuring their fatigue life across varied production environments \cite{malek2020fatigue}. The stress-life diagram, commonly known as the S-N curve, graphically represents the relationship between applied stress amplitude and the number of cycles until material failure \cite{burhan2018sn}. This serves as a fundamental method for characterizing fatigue behavior in materials. S-N curves play an essential role in evaluating safety and durability characteristics. Unfortunately, conducting numerous fatigue experiments to generate these curves is both financially burdensome and extraordinarily time-consuming, particularly for high-cycle fatigue testing. Therefore, developing predictive models for S-N curves represents a valuable approach that can efficiently support both the characterization of fatigue properties and the strategic development of alloys engineered for specific fatigue performance requirements.

It is very challenging to develop a universal model for predicting S-N curves due to fatigue's inherent complexity and limited experimental data. Instead, researchers have created various empirical fatigue models that transform experimental data into mathematical equations approximating S-N curve behavior, such as Basquin’s exponential law~\cite{oh1910exponential} and Manson–Coffin relationship~\cite{manson1953behavior}. Additionally, the Stüssi function models a nonlinear S-N curve with fatigue-limited considerations to improve the flexibility of fatigue life predictions~\cite{caiza2018probabilistic}. The Weibull distribution, originally developed as a general statistical model, has been widely adopted to describe fatigue failure probability and model material reliability under varying stress conditions~\cite{weibull1951statistical}. The Pascual and Meeker (PM) model estimates fatigue strength using a statistical framework that accounts for random fatigue limits and experimental data variability to enhance fatigue life assessment~\cite{pascual1999estimating}. However, these empirical S-N models often fail as predictive tools due to parameter uncertainties and material-specific variations in curve shapes. While they effectively fit data for individual materials, they typically cannot accurately predict fatigue behavior across different materials~\cite{morrow1965cyclic,lagoda2022using}.


Machine learning (ML) and deep learning (DL) have emerged as powerful alternatives for material science~\cite{decost2020scientific, li2020ai, liu2022material} due to their prediction power.  Traditional ML models, including Random Forests, Support Vector Machines, and XGBoost, have been applied to capture complex nonlinear relationships within fatigue data for various materials and conditions~\cite{agrawal2014exploration,bao2021machine,guo2023random}. Further, Dai et al. \cite{dai2020using} used artificial neural networks to predict the fatigue life of laser powder bed fusion stainless steel. Wei et al. \cite{wei2022high} employed the long short-term memory (LSTM) network combined with the transfer learning concept to estimate the high cycle S-N curve.  Zhu et al. \cite{zhu2024high} proposed Multi-Graph Attention Networks (Multi-GAT) to predict the high cycle fatigue life of several types of titanium alloys. 

Despite the success of ML/DL models, one challenge is that the size of the fatigue experimental dataset is usually small, which raises concerns of ML/DL about poor extrapolation beyond training conditions~\cite{xu2023small}. Researchers have proposed to incorporate domain knowledge into ML and DL models.  Lian et al.~\cite{lian2022fatigue} successfully combined empirical knowledge with ML for aluminum alloy fatigue life prediction, achieving significantly improved accuracy. Gan et al.~\cite{gan2023integration} incorporated not only real data from experiments but also dummy data generated by empirical fatigue life models to augment training datasets. Chen et al. \cite{chen2024crack} implemented the physics-informed neural networks (PINNs) \cite{raissi2019physics} for the fatigue crack growth simulation and life prediction of structures. However, these models are restricted to traditional ML/simple neural networks with limited domain knowledge. 

To address these challenges, we introduce DeepOFormer, a deep operator learning with domain-informed features, for fatigue life prediction of aluminum alloys. Specifically, the contributions of our paper can be summarized as follows.
\begin{itemize}
    \item We formulate the S-N curve prediction problem as an operator learning problem and propose a novel model called DeepOFormer.
    \item We explored multiple domain-informed features such as Stüssi, Weibull, and PM features.
    \item We use data from \cite{lian2022fatigue} to demonstrate the effectiveness and efficiency of DeepOFormer by comparing it with state-of-the-art deep learning methods.
\end{itemize}

The rest of this paper is organized as follows. The Dataset used in this paper is introduced in Section~\ref{sec: data}, followed by
our proposed DeepOFormer in Section~\ref{sec: method}. Our experiments and
results are summarized in Section~\ref{sec: exp}. Finally, conclusions are
discussed in Section~\ref{sec: conclusion}.
\section{Dataset Description}\label{sec: data}

The S-N curve can illustrate the relationship between stress amplitude ($\sigma_a$) applied during cyclic loading and the number of cycles ($N$) to failure. It is typically plotted with stress amplitude on the vertical axis and the number of cycles on the horizontal axis. It provides a graphical representation essential for predicting fatigue performance and establishing safe operating limits. The fatigue S-N curve dataset used in this work is from \cite{lian2022fatigue}, which has 54 S-N curves from seven types of aluminum alloys comprising 257 high-cycle (10\textsuperscript{4}--10\textsuperscript{10} cycles) tensile fatigue test data.  All tests are performed at room temperature conditions. 

In the dataset, mechanical properties such as
ultimate tensile stress (UTS), tensile yield strength (TYS), and Fatigue Strength are taken into consideration because of their high positive correlation with fatigue life. Temper \cite{cayless1990alloy} is one important categorical feature that contains information on treatments in aluminum alloy production and correlates to the microscopic structure. Stress ratio (R) that equals $\sigma_{max}/\sigma_{min}$ is also included in the feature set, where $\sigma_{min}$ is minimum cyclic stress, $\sigma_{max}$ is maximum cyclic stress. We also consider stress amplitude $\sigma_a$ and $\sigma_a^3$ in the feature set for modeling.  The logarithmic of fatigue life, i.e., $\log(N)$, is used as the target variable in the dataset for modeling. 



In this paper, we also consider multiple domain-informed features, which are motivated by empirical fatigue models. The Stüssi feature \cite{caiza2018probabilistic} normalizes the UTS relative to a reference stress amplitude level $\sigma_a$ and the Fatigue Strength. Formally, we have the following
\begin{equation}\label{eq:Stüssi}
    \text{Stüssi} 
    \;=\; \log \left( 
        \frac{\text{UTS} - \sigma_a}{\sigma_a - \text{Fatigue Strength} + 100}
    \right),
\end{equation}
where $100$ serves as a positive constant to maintain a positive quantity in the denominator. By comparing $\text{UTS} - \sigma_a$ to $\sigma_a - \text{Fatigue Strength}$, this feature highlights how far the applied or reference stress is from both the material’s ultimate tensile strength and its fatigue limit.

The Weibull feature \cite{weibull1951statistical} is introduced to capture statistical failure behavior by quantifying the relative deviation of applied stress from the material’s threshold. It is a statistical approach to predict material failure by modeling the probability of survival based on stress variations. Specifically, we have
\begin{equation}
        \label{eq:weibull}
        \text{Weibull} = 
        \log \left( \frac{\text{UTS} - \text{Fatigue Strength}}{\sigma_a - \text{Fatigue Strength} + 100} \right).
\end{equation}
The PM feature \cite{pascual1999estimating} normalizes fatigue strength with respect to a shifted stress difference as
\begin{equation}
    \label{eq:pm}
    \text{PM } = 
    \log \left( \frac{\text{Fatigue Strength}}{\sigma_a - \text{Fatigue Strength} + 100} \right).
\end{equation}

\begin{figure}[htbp]
  \centering
  \includegraphics[width=0.5\textwidth]{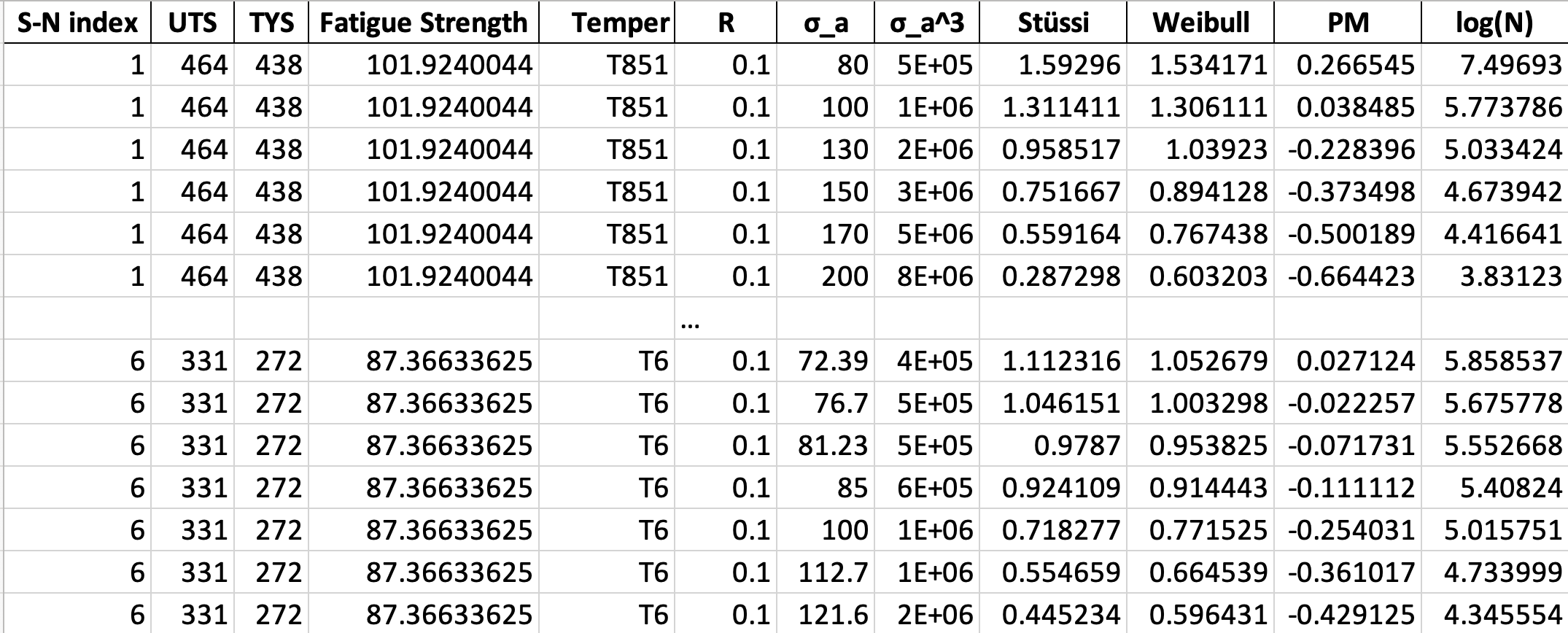} \vspace{-0.5cm}
  \caption{Screenshot of the fatigue dataset used in this paper.}
  \label{fig:data}
\end{figure}
Fig.~\ref{fig:data} shows the screenshot of the fatigue dataset. The first column shows the S-N curve index, where we have 54 curves in total. The second to sixth columns show the UTS, TYS, Fatigue Strength, Temper, and R, respectively. These features remain constant for different data points for a specific S-N curve. Since UTS, TYS, and Fatigue Strength are mechanical properties of the material. Temper is a categorical variable for material processing and R is the experimental setting. The seventh to eleventh corresponds to $\sigma_a$, $\sigma_a^3$, Stüssi, Weibull, and PM features. The last column is the target variable $\log(N)$.



\begin{figure*}[htbp]
  \centering
  \includegraphics[width=0.95\textwidth]{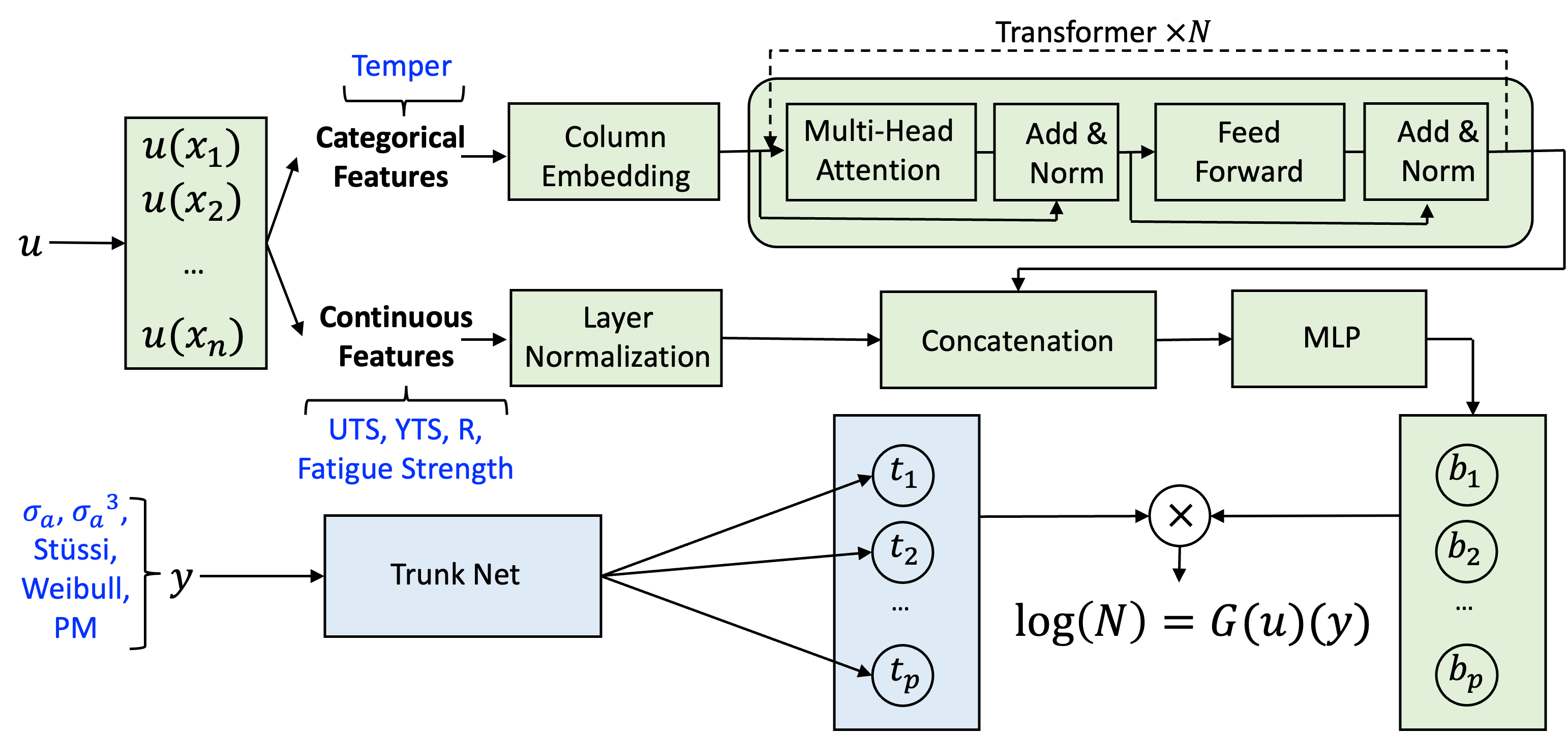}
 \caption{DeepOFormer architecture for fatigue life prediction. A transformer network (light green) encodes features $u=$ [UTS, TYS, Fatigue Strength, Temper, R], while the trunk network (light blue) processes stress-related and domain-informed features $y=$ [$\sigma_a$, $\sigma_a^3$, St\"ussi, Weibull, PM]. The final $\log(N)$ is obtained via the operator-based dot product of the two embeddings. }
  \label{fig:structure}
\end{figure*}

\section{METHODOLOGY}\label{sec: method}
\subsection{Problem Formulation}
In this subsection, we will introduce how we formulate fatigue life prediction as an operator learning problem \cite{chen1995universal,lu2019deeponet}. Following the framework and notations in DeepOnet \cite{lu2019deeponet}, let $G$ be an operator taking an input function $u$, and then $G(u)$ is the corresponding output function. For any point $y$ in the domain of $G(u)$, the output $G(u)(y)$ is a real number. Hence, the network takes inputs composed of two parts: $u$ and $y$, and outputs $G(u)(y)$. Although our goal is to learn operators, which take a function as the input, we have to represent the input functions discretely, so that network approximations can be applied. A straightforward and simple way, in practice, is to employ the function values at sufficient but finite many locations $\{x_1, x_2, . . . , x_m\}$; we call these locations “sensors”. As shown in Fig.~\ref{fig:structure}, these “sensors” correspond to the feature set [UTS, TYS, Fatigue Strength, Temper, R] in our problem. $y$ is a vector with five components, i.e., [$\sigma_a$, $\sigma_a^3$, Stüssi, Weibull, PM]. 

By formulating the fatigue life prediction problem as an operator learning problem, we develop a Transformer-based operator learning framework called DeepOFormer. The details architecture of DeepOFormer is described in Section~\ref{subsec: DeepOFormer}.

\subsection{DeepOFormer Architecture}\label{subsec: DeepOFormer}

The proposed architecture, DeepOFormer, improves the deep operator learning framework with a Transformer-based encoder. As shown in Fig.~\ref{fig:structure}, our DeepOFormer architecture comprises two branches of networks: the upper branch is the Transformer-based encoder (depicted in light green) and the bottom branch is the trunk network (depicted in light blue).

\noindent  \textbf{Transformer-based Encoder.} The Transformer-based encoder is motivated from \cite{huang2020tabtransformer}. It can process both categorical features (e.g., Temper: T851, T6, T8, etc.) and continuous features (UTS, YTS, R, and Fatigue Strength). The categorical features first pass through a column embedding layer. A stack of $N$ Transformer layers transforms the embeddings of categorical features into contextual embeddings. Each Transformer contains multi-head self-attention blocks~\cite{vaswani2017attention}:
\begin{equation}
    \mathrm{Attention}(Q, K, V) 
    \;=\; \mathrm{softmax}\!\Bigl(\frac{Q K^{\top}}{\sqrt{k}}\Bigr)\, V,
\end{equation}
where $Q, K, V$ represent the query, key, and value matrices derived from input embeddings and $k$ is the projection dimension. For each embedding, the attention matrix $\mathrm{softmax}\!\Bigl(\frac{Q K^{\top}}{\sqrt{k}}\Bigr)$ calculates how much it attends to other embeddings, thus transforming the embedding into contextual one. The output of the attention head of dimension is projected back to the embedding of dimension through a fully connected layer, which in turn is passed through two position-wise feed-forward layers. The first layer expands the embedding to four times its size and the second layer projects it back to its original size.

The continuous features pass through a layer normalization and concatenate with the contextual embeddings from Transformers. The concatenated features pass through a Multilayer perception (MLP) to get the embedding $[b_1, b_2, \dots, b_p]^{\top}$ of dimension $p$.


\noindent  \textbf{Trunk Network.} The trunk network processes stress-related and domain-informed features $y=$ [$\sigma_a$, $\sigma_a^3$, St\"ussi, Weibull, PM]. The trunk network is a simple MLP with multiple hidden layers, outputting an embedding $[t_1, t_2, \dots, t_p]^{\top}$ of the same dimension $p$. The trunk network also applies activation functions in the last layer, i.e., $t_k = \sigma(\cdot)$ for $k = 1, 2, . . . , p$. 


\noindent  \textbf{Fatigue Prediction.} After embeddings from the Transformer-baed encoder and trunk network are obtained, the model predicts the fatigue life, i.e., $\log(N)$, by taking the element-wise product of the two embeddings:
\begin{equation} \label{eq:deeponet_operator}
    G(u)(y) \;=\; \sum_{i=1}^{p} b_i\,t_i \;+\; b_0,
\end{equation}
where the scalar $b_0$ is the bias term. Adding bias may increase the performance by reducing the generalization error. 

\noindent  \textbf{Loss Function.} We adapt a mean L2 relative error (ML2RE) loss:
\begin{equation}
    \ell_\mathrm{ML2RE} = \frac{1}{n}\sum_{j=1}^{n} \frac{(y_j - \hat{y}_j)^2}{y_j^2 + \epsilon},
\end{equation}
where $y_j$ is the true $\log(N)$, $\hat{y}_j$ is the model prediction, and $\epsilon$ is a small constant ($10^{-30}$) to prevent division by zero.

\noindent \textbf{Advantages of DeepOFormer.}  By adopting an operator learning perspective, our DeepOFormer can capture how changes in material-related features modulate the functional mapping from stress-related features to fatigue life. Consequently, DeepOFormer generalizes more robustly to new or unseen combinations of material-related features than conventional neural network approaches. Moreover, incorporating both the Transformer-based encoder and domain-informed features ensures data efficiency since our DeepOFormer leverages domain knowledge alongside advanced deep learning architecture. Overall, our proposed DeepOFormer is a scalable framework for accurate and interpretable fatigue life prediction across diverse aluminum alloys.

\section{EXPERIMENTS} \label{sec: exp}

\subsection{Experimental Setup}
Our dataset consists of 54 fatigue S-N curves, comprising a total of 257 data points. Each S-N curve contains a varying number of data points representing fatigue life under different stress amplitudes. For example, one S-N curve has five data points while another one has four data points. To evaluate the model’s performance, the dataset is partitioned into training and test sets at the curve level. Specifically, 47 curves (216 data points) are randomly selected for training. The remaining seven curves (41 data points) are selected exclusively for testing. These seven curves represent for seven different aluminum alloys.

In our DeepOFormer, the trunk network is a two-layer MLP with 64 hidden units per layer, employing Leaky ReLU \cite{xu2015empirical} as the activation function.  Meanwhile, the Transformer-based encoder contains two Transformer blocks, where each Transformer block contains one three-head attention with a dimension of 48 in each head. The dropout levels are set to 0.2 for attention layers and 0.1 for feed-forward layers. $p$ is set to 16 when combining embeddings from the Transformer encoder and the trunk network. 

All experiments are conducted in Python 3.9 using PyTorch-based DeepXDE on an NVIDIA A100 GPU. Each training run is repeated ten different times to assess model reliability. All deep learning models are optimized via the Adam optimizer \cite{kingma2014adam} where a learning rate of 0.001, a batch size of 64, and a maximum of 3000 training epochs. 

Three metrics are used for performance evaluation: coefficient of determination ($R^2$), mean absolute error (MAE), and mean relative error (MRE). For a dataset of size $n$, denoting $y_i$ as the true value of $\log(N)$, and $\hat{y}_i$ as the predicted value, we have 
\begin{equation}
    R^2 = 1 - \frac{\sum_{i=1}^n (y_i - \hat{y}_i)^2}{\sum_{i=1}^n (y_i - \overline{y})^2},
\end{equation}
where $\overline{y}$ is the mean of all $y_i$ and $\varepsilon=10^{-12}$ is a small constant that prevents division by zero. MAE and MRE are defined as follows
\begin{equation}
    \text{MAE} = \frac{1}{n} \sum_{i=1}^n \bigl|\, y_i - \hat{y}_i \bigr|,
\end{equation}
\begin{equation}
        \mathrm{MRE} = \frac{1}{n} \sum_{i=1}^n \frac{\bigl|\,10^{y_i} - 10^{\hat{y}_i}\bigr|}{\bigl|10^{y_i}\bigr| + \varepsilon}.
    \end{equation}
Note that in our experiment, the MRE is calculated after exponentiating both the true and predicted values, which are in the logarithm scale. By transforming $\log(N)$ back to the original domain $N$ (i.e., $10^{y}$), the metric effectively captures the error in terms of actual fatigue life rather than the logarithm scale.

\begin{figure*}[!htbp]
  \centering
  \includegraphics[width=\textwidth]{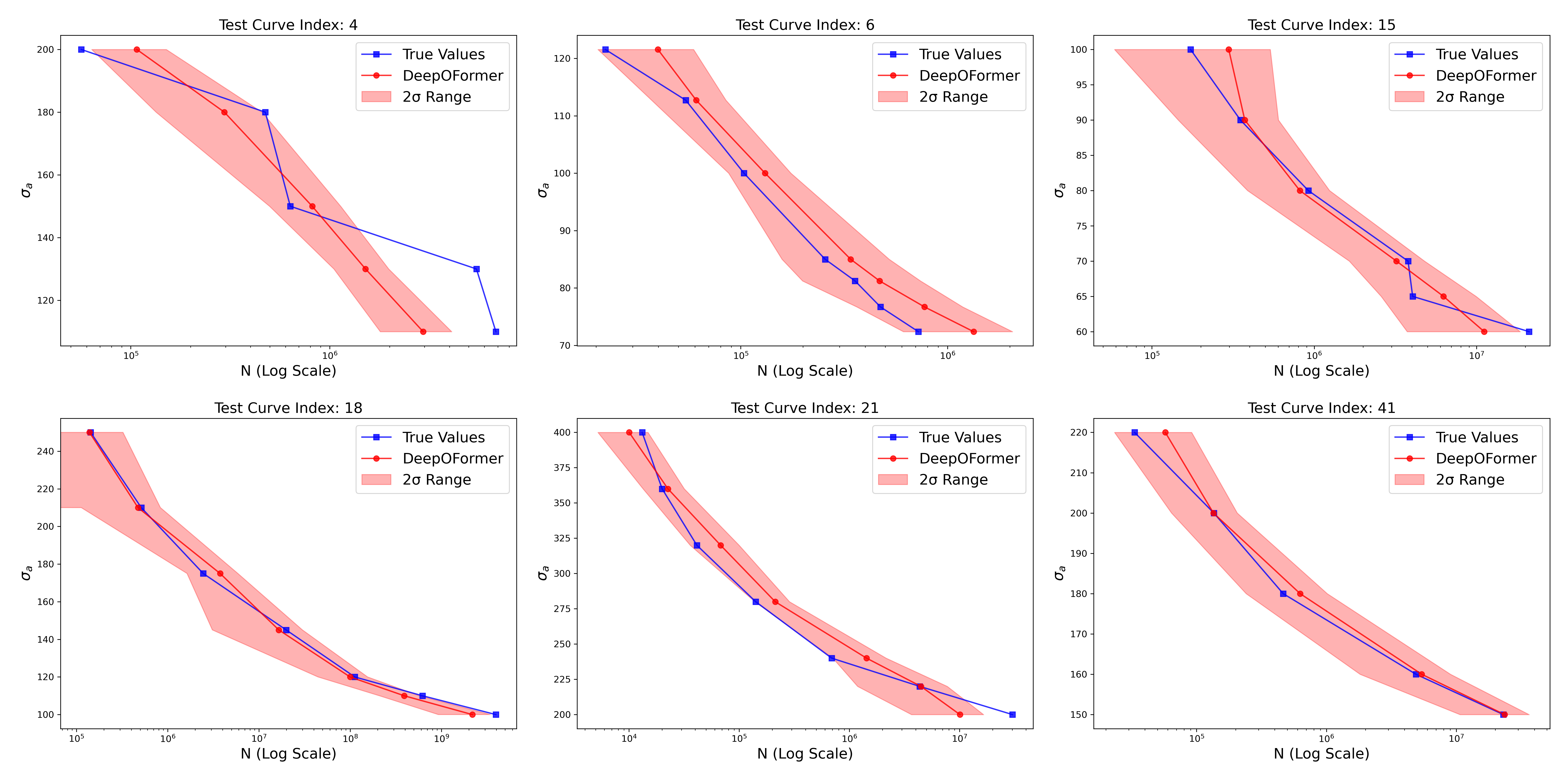}\vspace{-0.4cm}
  \caption{Predicted S-N curves (red circles) versus the true S-N curves (blue squares) for six selected S-N curves.  The shaded region denotes $\pm 2\sigma$ from ten repetitions, illustrating the model's uncertainty. 
    The $x$-axis shows the number of cycles ($N$) in the logarithm scale, and the $y$-axis is the stress amplitude $\sigma_a$. }
  \label{fig:curve_result}
\end{figure*}

\subsection{Results and Discussions}
In our experiments, we implement the following methods together with DeepOFormer for comparison. We have (1) DeepOFormer W/O ML2RE, where we use the Mean Squared Error (MSE) as a loss function; (2) DeepOFormer W/O Domain-informed Features; (3) DeepONet \cite{lu2019deeponet}; (4) DeepONet+Transformer: the branch network is replaced with Transformer \cite{vaswani2017attention};  (5) TabTransformer \cite{huang2020tabtransformer}; (6) XGBoost: the best methods identified in \cite{chen2016xgboost}.

\begin{table}[!htbp]
\centering
\caption{Average Performance of Ten Repetitions for Different Methods in terms of $R^2$, MAE, and MRE.}
\label{tab:results}
\begin{tabular}{lccc}
\hline
\textbf{Model} & \textbf{$R^2$ $\uparrow$} & \textbf{MAE$\downarrow$} & \textbf{MRE$\downarrow$} \\ \hline
DeepOFormer & \textbf{0.9515} & \textbf{0.2080} & \textbf{0.5077} \\
DeepOFormer W/O ML2RE & 0.9451 & 0.2184 & 0.5378 \\
DeepOFormer W/O        & \multirow{2}{*}{0.9339} & \multirow{2}{*}{0.2418} & \multirow{2}{*}{0.6235} \\
Domain-informed Features                                            &                   &                   &                                 \\
DeepONet \cite{lu2019deeponet} & 0.8200 & 0.3800 & 1.0136 \\
DeepONet+Transformer & 0.8700 & 0.3100 & 0.6200 \\
TabTransformer \cite{huang2020tabtransformer} & 0.9218 & 0.2548 & 0.6668 \\
XGBoost \cite{chen2016xgboost} & 0.8184 & 0.3854 & 1.0236 \\
\hline
\end{tabular}
\end{table}

\begin{figure}[!htbp]
  \centering
  \includegraphics[width=0.5\textwidth]{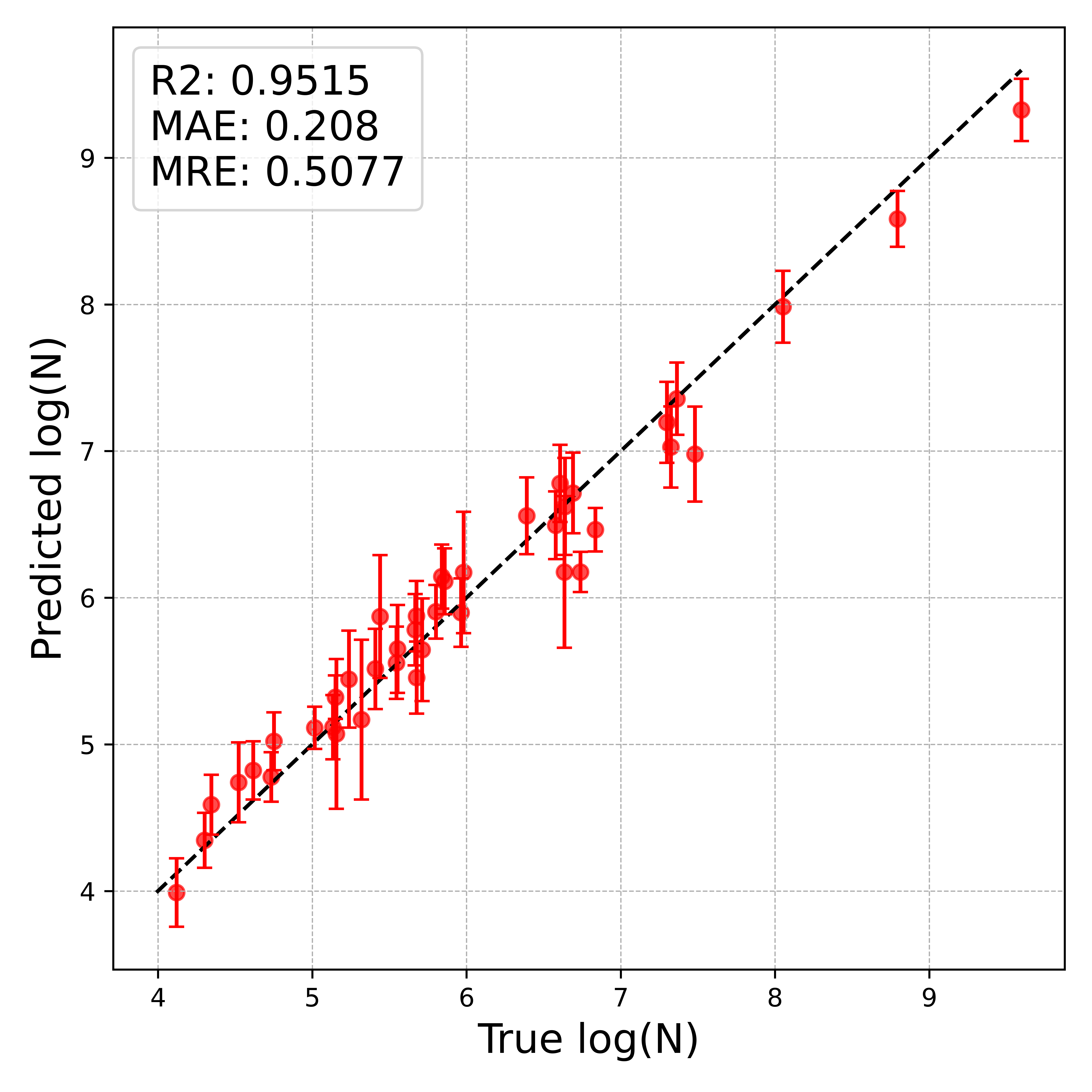} \vspace{-0.6cm}
  \caption{Predicted vs. true $\log(N)$ on the test set, with error bars showing $\pm 2\sigma$ from ten repetitions.  The diagonal line indicates perfect prediction.}\vspace{-0.2cm}
  \label{fig:$R^2$_result}
\end{figure}

The average prediction results of ten repetitions for different methods are summarized in Table~\ref{tab:results}.  XGBoost yields an $R^2$ of 0.8184, an MAE of 0.3854, and an MRE of 1.0236, which perform worse than all other deep learning methods. Among all deep learning methods, the proposed DeepOFormer model, which uses all three domain-informed features and the ML2RE loss function, significantly outperformed all other methods, achieving the highest performance with an $R^2$ of 0.9515, an MAE of 0.2080, and an MRE of 0.5077.  When applying the MSE loss, the performance of DeepOFormer W/O ML2RE slightly decreases to an $R^2$ of 0.9451, an MAE of 0.2184, and an MRE of 0.5378. This shows the importance of the ML2RE loss function. Without domain-informed features,  DeepOFormer W/O Domain-informed Features has $R^2$ = 0.9339, MAE = 0.2418, and MRE = 0.6235, which further demonstrates the importance of domain-informed features. DeepONet integrated with the Transformer architecture (DeepONet+Transformer) achieves moderate prediction accuracy with $R^2$ = 0.8700, MAE = 0.3100, and MRE = 0.6200. Additionally, the TabTransformer exhibits competitive performance, where $R^2$ = 0.9218, MAE = 0.2548, and MRE = 0.6668. The DeepONet has the lowest prediction accuracy among all the deep learning methods with $R^2$ = 0.8200, MAE = 0.3800, and MRE = 1.0136.  These results highlight that integrating our Transfromer-based encoder, domain-informed features, and a customized loss function can substantially improve the prediction and generalization of fatigue life prediction in aluminum alloys.

Fig.~\ref{fig:curve_result} provides a detailed visualization of fatigue life predictions across six selected testing curves (due to the limited space, we can only show six instead of seven), each corresponding to a different aluminum alloy. Specifically, the selected test curves with indices 4, 6, 15, 18, 21, and 41 correspond to six aluminum alloys 2618-T851, A357-T6 alloys, A6N01S-T5-A, 2024-T351m, A7075-T6-A, and A5083P-O-D, respectively. These curves are selected to compare the prediction performance of DeepOFormer with the ground truth. The blue line represents the true values, the red line represents DeepOFormer’s predictions, and the red-shaded region indicates the $\pm 2\sigma$ uncertainty range from ten different repetitions, where $\sigma$ is the standard deviation. As the fatigue life $N$ shown on a logarithm scale along the x-axis increases, both the model predictions and the ground truth generally follow a similar downward trend. Moreover, in most cases, the true values remain within the $\pm 2\sigma$ uncertainty interval of DeepOFormer’s predictions, suggesting that it effectively captures changes in the target quantity $\log(N)$ across different stress amplitudes $\sigma_a$. 

Fig.~\ref{fig:$R^2$_result} illustrates the predicted fatigue life ($\log N$) plotted against the true values for the test S-N curves. Each red point denotes the average prediction across ten repetitions. The red vertical intervals represent the corresponding $\pm 2\sigma$ intervals, capturing the standard deviation of ten repetitions. This plot demonstrates a strong correlation between predicted and true $\log(N)$ values, with most predictions falling close to the ideal diagonal line. The error bars showing $\pm 2\sigma$ from ten repetitions indicate reasonable uncertainty ranges, with wider uncertainty at some points but generally tight confidence intervals across the prediction range. In addition, most error bars can cover the perfect fitting line (diagonal line). This result demonstrates the model's reliability in accurately predicting fatigue life across diverse aluminum alloys in the test set. 

%

\section{CONCLUSION} \label{sec: conclusion}
This paper proposes, DeepOFormer, a deep operator learning model for fatigue life prediction of aluminum alloys. DeepOFormer has a Transformer encoder that considers categorical features and a customized loss function. We further improve the learning efficiency of DeepOFormer by considering domain-informed features such as Stüssi, Weibull, and PM features.  The prediction results of seven different aluminum alloys demonstrate the superior performance of DeepOFormer compared to traditional machine learning and deep learning models.

Despite the promising results achieved by the proposed DeepOFormer, there are still some limitations. The dataset used for training is relatively small, restricting the diversity of alloy compositions.  Our future work will focus on enriching the dataset. Additionally, designing a physics-informed loss function could provide stronger constraints on the learning process by penalizing physically implausible predictions. These refinements can improve DeepOFormer into a more general and reliable tool for fatigue life estimation, extending its utility to diverse industrial and research applications.
\section*{Acknowledgement}
This research in this paper is a collaboration with the U.S. Army Combat Capabilities Development Command Data \& Analysis Center (DEVCOM DAC), which provided essential technical expertise that significantly enhanced the scope and impact of our findings.


\bibliographystyle{IEEEtran}  
\bibliography{references}

\end{document}